\documentclass[draft]{ecai} 

\usepackage{bm}
\usepackage{latexsym}
\usepackage{amssymb}
\usepackage{amsmath}
\usepackage{amsthm}
\usepackage{booktabs}
\usepackage{enumitem}
\usepackage[final]{graphicx}
\usepackage{color}
\usepackage{algorithm}
\usepackage{enumitem}
\usepackage{algorithmicx}
\usepackage{algpseudocode}
\usepackage{amsfonts}
\usepackage[caption=false]{subfig}
\usepackage{tabularx}
\usepackage{multirow}
\usepackage{tcolorbox}
\usepackage{siunitx}
\usepackage[final]{listings}
\lstdefinelanguage{Prolog}{
    morekeywords={:-, ?, is, not, fail, true, listing, and, if},
    sensitive=true,
    morecomment=[l]{\%},
    morestring=[b]",
}

\definecolor{darkblue}{RGB}{0,0,139}

\newcommand{\gpt}{GPT-4o}
\newcommand{\gemini}{Gemini 1.0 Pro}

\begin{document}


\begin{frontmatter}

\title{Towards Logically Sound Natural Language Reasoning with Logic-Enhanced Language Model Agents}

\author[A]{\fnms{Agnieszka}~\snm{Mensfelt}\orcid{0000-0002-2385-2017}}
\author[A]{\fnms{Kostas}~\snm{Stathis}\orcid{0000-0002-9946-4037}}
\author[A]{\fnms{Vince}~\snm{Trencsenyi}\orcid{0009-0009-4560-7571}} 

\address[A]{Royal Holloway, University of London, Egham, Surrey, UK}


\begin{abstract}
Large language models (LLMs) are increasingly explored as general-purpose reasoners, particularly in agentic contexts. However, their outputs remain prone to mathematical and logical errors. This is especially challenging in open-ended tasks, where unstructured outputs lack explicit ground truth and may contain subtle inconsistencies. To address this issue, we propose Logic-Enhanced Language Model Agents (LELMA), a framework that integrates LLMs with formal logic to enable validation and refinement of natural language reasoning. LELMA comprises three components: an LLM-Reasoner, an LLM-Translator, and a Solver, and employs autoformalization to translate reasoning into logic representations, which are then used to assess logical validity. Using game-theoretic scenarios such as the Prisoner’s Dilemma as testbeds, we highlight the limitations of both less capable (\gemini{}) and advanced (\gpt{}) models in generating logically sound reasoning. LELMA achieves high accuracy in error detection and improves reasoning correctness via self-refinement, particularly in \gpt{}. The study also highlights challenges in autoformalization accuracy and in evaluation of inherently ambiguous open-ended reasoning tasks.
\end{abstract}

\end{frontmatter}


\section{Introduction}

Large language models (LLMs) are capable of generating sophisticated natural language explanations and decisions across a wide range of tasks. Their ability to process and produce human-like text makes them attractive candidates for use in agentic AI systems, where autonomous agents interact with both humans and other agents through natural language communication. This vision includes scenarios ranging from collaborative multi-agent systems to negotiation, coordination, and human-agent teaming~\cite{park2023generative, durante2024agent}.

However, despite their extensive capabilities, LLMs remain fundamentally unreliable reasoners. Their outputs often suffer from mathematical, logical, and factual errors, as well as hallucinations~\cite{imani2023mathprompter,achiam2023gpt}. Furthermore, although significant research effort has been devoted to improving single-agent safety and alignment, ensuring that a model produces safe and accurate outputs in isolation does not automatically guarantee safe and reliable behavior in multi-agent settings~\cite{anwar2024foundational}.

In this context, game theory~\cite{Osborne2004} offers a particularly suitable testbed. As a formal model of strategic interaction, it allows controlled evaluation of reasoning in multi-agent scenarios where choices must be justified based on payoffs, incentives, and anticipated actions of others. Recent works have explored LLMs as agents in game-theoretic environments, primarily focusing on action optimality -- whether models select payoff-maximizing strategies~\cite{akata2023playing,fan2023can,guo2023gpt,lore2023strategic}. However, optimality alone may result from well-represented game-theoretic examples in the training data, and thus does not guarantee robustness when the model encounters novel or out-of-distribution scenarios.

In this work, we shift the focus from action outcomes to reasoning processes. We introduce Logic-Enhanced Language Model Agents (LELMA), a framework that integrates LLMs with formal logic to validate and refine natural language reasoning. LELMA employs a modular architecture consisting of three components: an LLM-Reasoner, an LLM-Translator, and a formal Solver. Through a process of autoformalization~\cite{wu2022autoformalization,he2023solving,jiang2022draft, yang2023coupling,pan2023logic,cosler2023nl2spec,chen2023nl2tl,feng2023language}, the system translates natural language reasoning into formal logic queries, enabling logical validity checks. Detected errors trigger feedback, allowing the LLM to iteratively improve its reasoning through self-refinement.

Using classic game-theoretic scenarios such as the Prisoner's Dilemma as evaluation tasks, we investigate reasoning quality in both advanced (\gpt{}) and less capable (\gemini{}) models. Our results show that LELMA effectively improves reasoning correctness -- particularly for \gpt{}. Still, challenges remain in translation accuracy and evaluation of open-ended, ambiguous reasoning. This highlights the need for more robust methods to ensure that LLMs, when deployed as agents in interactive settings, reason logically soundly.

This paper makes the following contributions:
\begin{enumerate}[label=\textbf{(\roman*)}]
\item We introduce, a neurosymbolic framework that integrates LLM-generated reasoning with formal logic to validate and refine natural language outputs.
\item We conduct extensive experiments with \gpt{} and \gemini{}, showing improvements in reasoning correctness, especially for advanced models.
\item We identify key challenges, including limitations in autoformalization and evaluation of open-ended reasoning.
\end{enumerate}

The remainder of this paper is organized as follows. Section~\ref{sec:preliminaries} presents the preliminaries, covering large language models, game theory, and general game playing. Section~\ref{sec:lelma} introduces the LELMA framework, followed by the detailed methodology in Section~\ref{sec:methods}. Section~\ref{sec:results} presents the experimental evaluation and key findings. We discuss these results in depth in Section~\ref{sec:discussion}, and finally, Section~\ref{sec:conclusions} summarizes the work and outlines directions for future research.

\section{Preliminaries}
\label{sec:preliminaries}

\subsection{Large Language Models}

The term \textit{Large Language Model} typically refers to a model based on the transformer architecture with billions of parameters.
In this work, \gpt{} and \gemini{} were used to compare a state-of-the-art (SOTA) model with a less capable counterpart.

\paragraph{\gpt{}}
OpenAI's \gpt{}~\cite{gpt4o2024}---based on and successor of GPT-4 ~\cite{achiam2023gpt}---a current SOTA LLM. Its multi-modal architecture supports text, image and audio inputs and outputs. It was pre-trained with publicly accessible data and third-party resources, and fine-tuned with human feedback. \gpt{} demonstrates enhanced performance in reasoning tasks compared to its predecessor and is reported to have increased factuality. However, \gpt{} still faces challenges with reliability due to hallucinations.

\paragraph{\gemini{}}
\gemini{} is Google's previous largest model, with good general performance~\cite{geminiteam2024gemini}. The Gemini model family supports various forms of textual, visual and audio inputs and can produce interleaved textual and visual outputs. Gemini models are pre-trained on publicly available resources for each modality. They support external tool use and web search, increasing their performance in reasoning tasks.

\subsection{Game Theory}
\label{sec:game-theory}

Game theory provides the mathematical tools to analyse the strategic interaction between decision-makers \cite{Osborne2004}.  Its applications range from analysing day-to-day social situations to complex political or economic problems.

\subsubsection{Games in our experiments}

\begin{table}[H]
    \caption{General payoff matrix and specific payoff matrices for considered games.}
     \centering
    \label{tab:combined}

    \subfloat[\raggedright A general game in matrix form. \label{subtab:gen}]{
        \begin{tabular}{rcc}
        \toprule
        \textbf{row/col} & \textbf{~~~~~~L~~~~~~} & \textbf{~~~~~~~R~~~~~~~} \\
        \midrule
        \textbf{U}  & ($W_{row}$, $W_{col}$) & ($X_{row}$, $X_{col}$) \\
        \textbf{D}  & ($Y_{row}$, $Y_{col}$) & ($Z_{row}$, $Z_{col}$) \\
        \bottomrule
        \end{tabular}
    }

    \vspace{2ex}

    \subfloat[\raggedright Prisoner's Dilemma: $T>R>P>S$. \label{subtab:PD}]{
        \begin{tabular}{rcc}
        \toprule
        \textbf{row/col} & \textbf{~~Defect (D)~~} & \textbf{~~Coop. (C)~~} \\
        \midrule
        \textbf{Defect (D)} & $(1, 1)$ & $(5, 0)$ \\
        \textbf{Coop. (C)} & $(0, 5)$ & $(3, 3)$ \\
        \bottomrule
        \end{tabular}
    }

    \vspace{2ex}

    \subfloat[\raggedright Stag Hunt: $R>T>P>S$. \label{subtab:SH}]{
        \begin{tabular}{rcc}
        \toprule
        \textbf{row/col} & \textbf{~~~~Hare (D)~~~~} & \textbf{~~~~Stag (C)~~~~} \\
        \midrule
        \textbf{Hare (D)} & $(1, 1)$ & $(3, 0)$ \\
        \textbf{Stag (C)} & $(0, 3)$ & $(5, 5)$ \\
        \bottomrule
        \end{tabular}
    }

    \vspace{2ex}

    \subfloat[\raggedright Hawk-Dove: $T>R>S>P$. \label{subtab:HD}]{
        \begin{tabular}{rcc}
        \toprule
        \textbf{row/col} & \textbf{~~~Hawk (D)~~~} & \textbf{~~~Dove (C)~~~} \\
        \midrule
        \textbf{Hawk (D)} & $(0, 0)$ & $(5, 1)$ \\
        \textbf{Dove (C)} & $(1, 5)$ & $(3, 3)$ \\
        \bottomrule
        \end{tabular}
    }

\end{table}

We experiment with symmetric non-cooperative games presented in normal form, usually a matrix showing the players, strategies, and payoffs \cite{Rasmusen2004introtogametheory}. Table~\ref{subtab:gen} demonstrates a general game matrix, where the row player's actions are denoted by U and D, and the column player's actions are marked by L and R. The four possible outcomes W,X,Y,Z denote the players' payoff $\pi_i$ for the given action pairs as the tuple $(\pi_{row},\pi_{col})$.

All games in our experiment can be mapped over the general game matrix. We classify actions U and L as ``Defect'' denoted by $D$ and actions D and R as ``Cooperate'' denoted by $C$. This allows us to use the well-known terminology from Axelrod's tournaments to define these games in terms of the four outcomes:

\begin{description}
    \item[\ensuremath{\mathbf{T}}] (Temptation to defect): $(D, C)$
    \item[\ensuremath{\mathbf{R}}] (Reward for mutual cooperation): $(C, C)$
    \item[\ensuremath{\mathbf{P}}] (Punishment for mutual defection): $(D, D)$
    \item[\ensuremath{\mathbf{S}}] (Sucker's payoff): $(C, D)$
\end{description}

The Prisoner's Dilemma (PD) is a classical concept in game theory involving two players. Each player can decide to cooperate for mutual benefit or betray their partner (defect) for individual reward, characterised by the matrix in Table~\ref{subtab:PD}. The Stag Hunt (SH) is a coordination game, where the players have two options: be selfish and hunt Hare (D), or be cooperative and hunt Stag (C). The SH has a payoff structure that favours mutual cooperation the most, instead of the temptation to defect as in the PD---Table~\ref{subtab:SH}. Our third game, Hawk Dove (HD), involves the defective action Hawk, and cooperative action Dove. As demonstrated by the game matrix in Table~\ref{subtab:HD}, HD is described by having the temptation to defect as the most rewarding outcome, but the mutual defection yields the least amount of payoff.

\subsection{General game playing}

General game playing (\cite{gdl}) aims at creating intelligent systems that understand the rules of arbitrary new games and learns to play them without human intervention. The Game Description Language (GDL) has been proposed as a formal, machine-processable language for describing the rules of arbitrary games (\cite{gdl-orig}). GDL focused on information games only, so it was extended in GDL-II (\cite{gdl-ii}) to cover n-player games with incomplete information and games in extensive normal form (\cite{gdl-univ}). GDL-II is based on the standard syntax and semantics of logic programming and characterised by the special keywords shown in Table~\ref{tab:game_functions}.

\begin{table}[htb]
\caption{Subset of GDL-II (uppercase letters denote variables, lowercase letters denote predicates/function symbols)}
\begin{tabularx}{\columnwidth}{>{\ttfamily\small}lX}
\toprule
role(R)           & {\tt\small R} is a player         \\ 
init(F)           & {\tt\small F} holds in the initial position \\
true(F)           & {\tt\small F} holds in the current position \\ 
legal(R, M)       & {\tt\small R} can do move {\tt\small M} in the current position \\ 
does(R, M)        & player {\tt\small R} does move {\tt\small M}  \\
next(F)           & {\tt\small F} holds in the next position \\
terminal          & the current position is terminal \\
goal(R, N)        & {\tt\small R} gets {\tt\small N} points in the current position\\
\bottomrule
\end{tabularx}
\label{tab:game_functions}
\end{table}

A difficulty with GDL systems is that learning without human guidance poses a reasoning challenge. Players must infer the possible actions of others, essentially evaluating hypothetical game situations before taking action. Action formalisms like the classical Situation Calculus~\cite{sc} have been developed for precisely this purpose. Formal schemes and inference methods are readily available for Situation Calculus~\cite{sc-games1,sc-games2}, while their deployment in general game playing presupposes a  translation from GDL into existing, suitably expressive action languages. One such scheme~\cite{Schiffel_Thielscher_2011} shows how to fully embed GDL-II into a version of the Situation Calculus based on knowledge fluents~\cite{sc-kbf}. Our game solver, presented later in section~\ref{subsec:solver}, is inspired by these previous works, to support light GDL specifications that can be readily combined with LLM APIs to validate their reasoning.

\section{Logic-Enhanced Language Model Agents}
\label{sec:lelma}

\subsection{Framework Overview}

\begin{figure}[h]
    \centering
    \includegraphics[width=\linewidth]{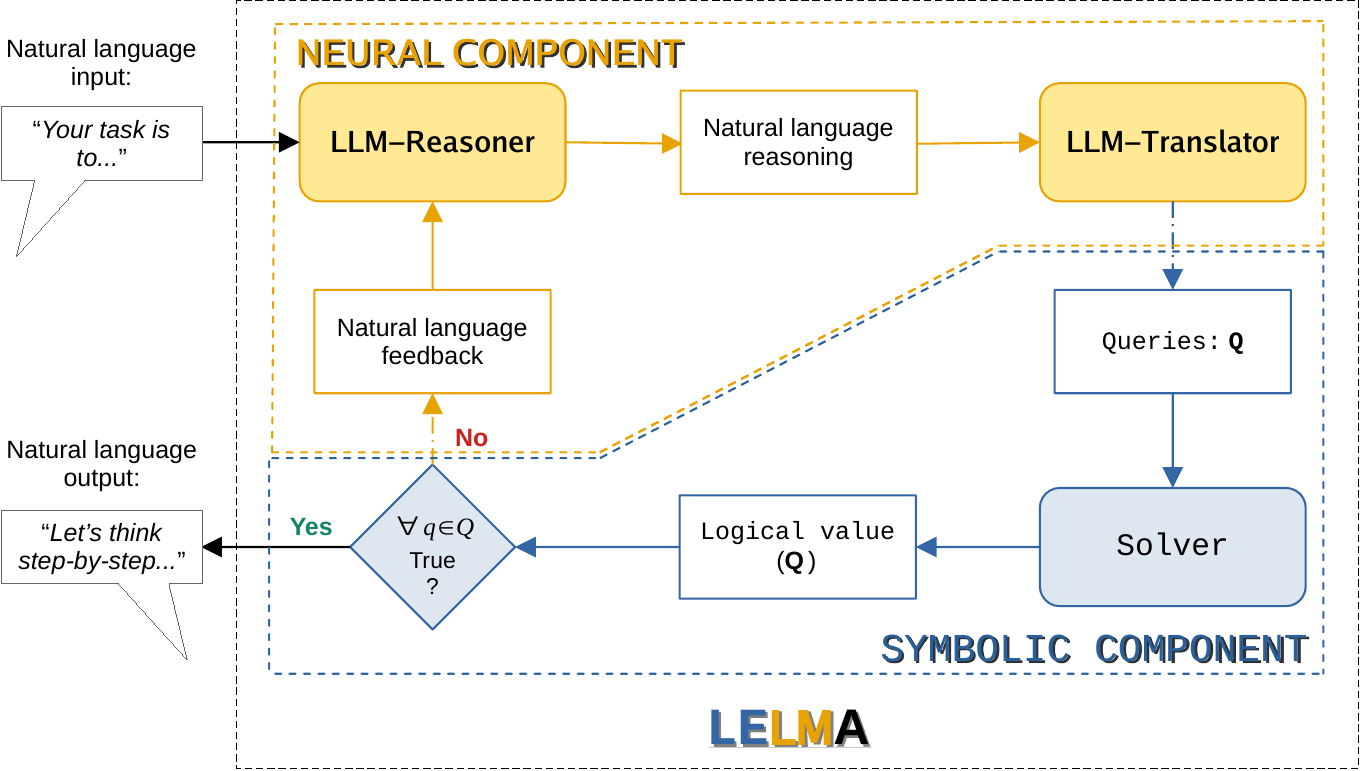}
    \caption{Overview of the LELMA framework. The relevant statements generated by the \textit{Reasoner} are translated into a set of queries \textbf{\texttt{Q}}, which are evaluated by the \textit{Solver}. If any of the queries fail, the \textit{Reasoner} is prompted to refine its reasoning in light of the facts derived from the failed queries. 
    }
    \label{fig:lelma}
\end{figure}

\label{sec:overview}
The Logic-Enhanced Language Model Agents (LELMA) framework is implemented in Python and Prolog~\cite{wielemaker2012swi}. Owing to the framework's modularity, the Prolog solver can be easily replaced with another type of formal solver. The system (Fig.~\ref{fig:lelma}) comprises the following components:
\begin{description}
\item[Reasoner:] An LLM agent responsible for generating natural language reasoning.
\item[Translator:] An LLM agent that maps statements from the \textit{Reasoner}'s output to logical queries sent to the \textit{Solver}.
\item[Solver:] A standard logic program implemented in Prolog that evaluates the queries.
\item[Feedback loop:] A mechanism for providing natural language feedback to the \textit{Reasoner} when a query evaluation fails.
\end{description}

\begin{algorithm}[htb]
\footnotesize
\caption{LELMA reasoning loop}
\label{alg:lelma-loop}
\textbf{Input}: Payoff matrix, instruction prompt\\
\textbf{Output}: Strategic reasoning
\begin{algorithmic}[1]
\State \texttt{attempts} $\gets 0$
\While{\texttt{attempts} $<$ \texttt{max\_attempts}}
    \State \texttt{reasoning} $\gets$ \texttt{Reasoner(instruction)}
    \State \texttt{queries} $\gets$ \texttt{Translator(reasoning)}
    \If{\texttt{queries} \textbf{is} $\varnothing$} 
        \State \Return \texttt{reasoning}  \Comment{\textcolor{blue!50}{\textit{No queries found}}}
    \Else
        \State \texttt{failed} $\gets$ \texttt{Solver(queries)}
        \If{\texttt{failed} \textbf{is} $\varnothing$} 
            \State \Return \texttt{reasoning} \Comment{\textcolor{blue!50}{\textit{All queries true}}}
        \Else
            \State \texttt{instruction} $\gets$ \texttt{feedback(failed)}
        \EndIf
    \EndIf
    \State \texttt{attempts} $\gets$ \texttt{attempts} $+ 1$ \Comment{\textcolor{blue!50}{\textit{Reevaluate using feedback}}}
\EndWhile
\State \Return \texttt{reasoning} \Comment{\textcolor{blue!50}{\textit{Max attempts reached}}}
\end{algorithmic}
\end{algorithm}

The source code of the framework and the results of the experimental evaluation are publicly available.\footnote{\textcolor{blue}{\url{https://github.com/dicelab-rhul/LELMA}}}

\subsection{Solver}
\label{subsec:solver}

We specify the solver module as a logic program implemented in Prolog~(\cite{p50}, \cite{wielemaker2012swi}), a declarative programming language well-suited for high-level problem definitions. Logic programs enhance the reliability of stochastic LLM outputs by providing a robust framework for logical rule-based reasoning. Our solver consists of three main components: a game-independent part representing the rules of any game in normal form, a game-dependent part defining the rules of a specific game using the predicates of the game-independent part, and a set of auxiliary predicates built on top of these representations to evaluate LLM reasoning for that specific game. We adhere to the usual conventions for representing a game as an implementable logic program: variables are denoted by uppercase letters and predicates and function symbols by lowercase letters. An underscore (\texttt{\_}) denotes a variable whose value is unused within a definition. The state of a game is represented by a situation, initially defined as a constant (e.g., \texttt{s0}). The special binary function \texttt{do(M, S)} represents the situation resulting from the execution of move \texttt{M} in situation \texttt{S}, as in the Situation Calculus. In the logical rules that follow, the symbols \texttt{if}, \texttt{and}, and \texttt{not} correspond to their logical equivalents. 

\subsubsection{Game-independent part}
We recursively specify as a logic program all legal transitions of an extended form game from an initial situation {\tt S} to a final situation {\tt F} as follows: 

\begin{lstlisting}
game(F, F) if 
    final(F).
game(S, F) if 
    not final(S) and 
    legal(M, S) and 
    game(do(M, S), F).
\end{lstlisting}
The first two lines return the final game situation {\tt F}, when it is reached. Otherwise, in a non-final situation {\tt S}, the game accepts a legal move {\tt M}, and the game continues in the next {\tt do(M,S)} situation, until the final situation {\tt F} is reached. To reason about what holds in individual legal situations, we use Situation Calculus represented as:

\begin{lstlisting}
holds(F, S) if 
    initially(F, S).
holds(F, do(M, S)) if 
    effect(F, M, S).
holds(F, do(M, S)) if 
    holds(F, S) and 
    not abnormal(F, M, S).
\end{lstlisting}

A fluent {\tt F} holds in the initial situation, a new fluent {\tt F} is initiated by the effects of a move {\tt M} executed in a situation {\tt S}, and a fluent {\tt F} persists after a move is made, provided it is not abnormal; abnormal fluents are terminated (do not persist). We also use rules of the form:
\begin{lstlisting}
finally(F, S) if Conditions.
\end{lstlisting}
to return derived fluents {\tt F} describing the result of the game, when the {\tt Conditions} hold  in the final situation {\tt S}. 

\subsubsection{Game-dependent part} To represent a specific game we need to define game-dependent predicates for the initial state {\tt initial/1}, the legal moves {\tt legal/2}, what holds in the initial game situation via {\tt initially/2}, the effects of a move on a situation via {\tt effect/3}, what stops persisting in a situation after the execution of a move via {\tt abnormal/3}, the final situation {\tt final/1}, and the result of the game via {\tt finally/2} definitions. To exemplify these definitions, we show how to describe a PD game to our solver.
The initial situation {\tt s0} is defined as:

\begin{lstlisting}
initial(s0).
\end{lstlisting}
What holds in this initial situation we specify it as:
\begin{lstlisting}
initially(player(p1), s0).
initially(player(p2), s0).
initially(role(p1,row), s0).
initially(role(p2,col), s0).
initially(control(p1), s0).
initially(control(p2), s0).
\end{lstlisting}
These assertions define first the player names represented by unique identifiers ({\tt p1} and {\tt p2}), their roles ({\tt p1} is the {\tt row} player, while {\tt p2} is the {\tt col}umn player), and then the fact that initially either of them can the game next (via the {\tt control/1} fluent,as in GDL). What holds in the initial situation changes as a result of move being made in the game. We represent moves as terms of the form {\tt choice(P, M)}, where {\tt P} is a player, and {\tt M} is a move. As it is possible for any player in a Prisoner's Dilemma game to choose defect ({\tt 'D'}) or cooperate ({\tt 'C'}), we write this as:
\begin{lstlisting}
possible(choice(P, 'D'), S) if 
    holds(player(P), S).
possible(choice(P, 'C'), S) if 
    holds(player(P), S).
\end{lstlisting}
It is then legal for a player to choose a possible move if they have the control to execute it in the current situation:
\begin{lstlisting}
legal(choice(P, M), S) if 
    possible(choice(P, M), S) and 
    holds(control(P), S).
\end{lstlisting}
When a legal move {\tt M} is made by a player {\tt P}, the effect that this move is actually made is recorded in the next situation:
\begin{lstlisting}
effect(did(P, M), choice(P, M), S).
\end{lstlisting}
Once a legal move is executed, the player loses control, which we specify in our framework as:
\begin{lstlisting}
abnormal(control(P), choice(P, M), S).
\end{lstlisting}
In other words, after a choice is made by a player, the player loses control and therefore cannot play a move again from that situation onwards. 

Moves made in this way bring about the final situation, which we specify as a situation term with two choices made from the initial situation, one for each player.
\begin{lstlisting}
final(S) if 
    ground(S) and 
    S = do(choice(_, _), do(choice(_, _), I)) and 
    initial(I).
\end{lstlisting}
Assuming the payoff matrix of Table~\ref{subtab:PD} defined as:
\begin{lstlisting}
payoff('D', 'D', 1, 1).
payoff('C', 'D', 0, 5).
payoff('D', 'C', 5, 0).
payoff('C', 'C', 3, 3).
\end{lstlisting}
the outcome of the game holds information about the actual moves made by the players, and their payoffs.
\begin{lstlisting}
finally(outcome(P1,M1,U1,P2,M2,U2), S) if
    final(S) and
    holds(role(P1, row), S) and
    holds(did(P1, M1), S) and
    holds(role(P2, col), S) and
    holds(did(P2, M2), S) and
    payoff(M1, M2, U1, U2).
\end{lstlisting}
We can then extract specific outcome information, e.g. the player's utility, through a {\tt goal/2} fluent (as in GDL) e.g.:
\begin{lstlisting}
finally(goal(P1, U1), S) if 
    finally(outcome(P1, _, U1, _, _, _), S).
finally(goal(P2, U2), S) if 
    finally(outcome(_, _, _, P2, _, U2), S).
\end{lstlisting}
This completes the definition of a PD game, and allows us to use the above game description to reason about the game. If, in the specifications above, we substitute the symbol \texttt{if} with \texttt{`:-'}, the symbol \texttt{and} with \texttt{`,'}, and the symbol \texttt{not} with \texttt{`\textbackslash +'} (representing negation by failure), we obtain an executable Prolog program that a player can query to perform reasoning within the game. For example, if player \texttt{p1} wanted to reason about the game and determine the actions needed to achieve a utility of 5, they could express this as the query shown below (queries are initiated with the symbol \texttt{`?-'}):

{\small {\tt ?- game(s0,F), finally(goal(p1,5),F).}}

based on the game's payoff matrix, our solver provides the following answers for \texttt{F}:

{\small 1\ {\tt do(choice(p2,'C'),do(choice(p1,'D'),s0));}}
{\small 2\ {\tt do(choice(p1,'D'),do(choice(p2,'C'),s0));}}
{\small 3\ {\tt false.}}

In the first answer, {\tt p1} acted first and {\tt p2} second, while in the second answer, the order is reversed. This is not unexpected, as in our game definition both players have control initially, so the answers show both combinations.

\subsection{Translator}

\begin{table}[htb]
\renewcommand{\arraystretch}{1.2}
\footnotesize
\caption{A set of queries corresponding to natural language statements and their explanation, as provided to the \textit{Translator}. For each predicate that determines the higher/highest payoff of a given type, except for $^*$, there exists a corresponding predicate determining the lower/lowest payoff.}
\centering
\begin{tabular}{|p{0.95\columnwidth}|}
\hline
\texttt{finally(outcome(you,B,1,them,R,\$ \_),S)} \\
`you' corresponds to the reasoner, `them' to their opponent, 10 to payoff, and `B', `R' to choices assumed to give this payoff to the reasoner \\
\hline
\texttt{higher(1, 3)} \\
1 and 3 correspond to numerical payoffs, 1 is assumed to be higher than 3 \\
\hline
\texttt{highest\_possible\_individual\_payoff(1)} \\
1 corresponds to the assumed highest possible payoff for the reasoner \\
\hline
\texttt{highest\_individual\_payoff\_for\_choice(1,B)} \\
1 corresponds to the assumed highest individual payoff for a given choice, e.g. `B' \\
\hline
\texttt{highest\_guaranteed\_payoff\_choice(B)}$^*$ \\
`B' corresponds to the choice assumed to give the highest guaranteed (worst-case) individual payoff \\
\hline
\texttt{higher\_guaranteed\_payoff(B,R)} \\
`B' and `R' correspond to choices, and `B' is assumed to give the higher guaranteed payoff \\
\hline
\texttt{highest\_mutual\_payoff(R,R)} \\
`R' and `R' correspond to choices, assumed to give the highest mutual payoff \\
\hline
\end{tabular}
\label{tab:evaluated-predicates}
\end{table}

\textit{The Solver} is extended with predicates derived from a preliminary analysis of common LLM reasoning errors in the evaluation task. Translation employs one-shot learning, using a prompt that lists predicates with sample arguments and their natural language descriptions (Table~\ref{tab:evaluated-predicates}). For failed predicates, \textit{the Solver} identifies correct variable values; the queries are then translated back into natural language and incorporated into feedback for \textit{the Reasoner}. Queries that are syntactically invalid or do not match expected forms are discarded. Constants outside the allowable set are replaced with variables.

\subsection{Feedback loop}

The feedback loop logic is shown in Listing~\ref{alg:lelma-loop}. The \texttt{max attempts} parameter limits the number of reasoning iterations. If all attempts fail (i.e., each contains some failed queries), the final attempt is returned as the most refined. The feedback prompt is dynamically populated with natural language corrections for each failed query. For example, if the original reasoning includes ``I choose B because it gives me payoff 3 which is higher than 5,'' the corresponding query \texttt{higher(3, 5).} fails and is translated as ``payoff 3 is lower than payoff 5.'' The prompt instructs the agent to reconsider its reasoning in light of these corrections, but also allows reaffirming the original reasoning. This accounts for possible false positives---that is, queries that were incorrectly translated. While the \textit{Solver} guarantees the correctness of the generated feedback, some feedback—--such as that suggesting a particular choice yields a higher guaranteed payof---may potentially bias the agent's reasoning toward that option.

\section{Methods}
\label{sec:methods}

\subsection{Task}

To evaluate the effectiveness of the LELMA framework in detecting and correcting natural language reasoning errors, we conducted experiments using the games introduced in Section~\ref{sec:game-theory}. The rules and payoff structures of each game were presented in a natural language instruction prompt, following the style commonly used in human-subject game-theoretic studies. The expected output was an action choice accompanied by natural language reasoning, consistent with the Chain-of-Thought~\cite{wei2023chainofthought} approach. Correctness was defined as the absence of logical errors in the reasoning about payoffs. 

To evaluate both the robustness of the approach and the baseline performance of LLMs, we deliberately obfuscated the task—that is, we modified its presentation to differ from formats likely present in the training data. Prior work has shown that LLM performance can degrade when familiar tasks are presented in unfamiliar ways~\cite{kambhampati2024llms}. In our setup, the standard game-theoretic action labels were replaced with abstract symbols ‘B’ and ‘R’. Furthermore, preliminary experiments indicated improved performance from \gpt{} when the payoff matrix followed a specific layout—particularly when the Cooperate/Cooperate outcome appeared in the top-left corner. Consequently, we inverted the matrix order in the experimental instructions.

\begin{table}[htb]
\renewcommand{\arraystretch}{1.2}
    \centering
        \caption{Experimental parameters. $^*$\textit{gpt-4o-2024-05-13}.}
    \small
    \begin{tabular}{|l|l|}
    \hline
    \multirow{3}{*}{\texttt{Games}} & Hawk-Dove \\
                                    & Prisoner's Dilemma \\
                                    & Stag Hunt \\
     \hline
    \multirow{2}{*}{\texttt{LLM-Reasoner Agents}} & \gpt{}$^*$ \\
                                                  & \gemini{} \\ 
    \hline
    \texttt{Temperature} & $1$ \\ \hline
    \texttt{Maximum output tokens} & $1024$ \\ \hline
    \texttt{Maximum attempts number} & $5$ \\ \hline
    \texttt{Repetitions} & $30$ \\ \hline

    \end{tabular}
    \label{tab:exp-parameters}
\end{table}

\subsection{Settings}

In the experiments, we used two LLMs as \textit{LLM-Reasoner} agents: \gpt{} and \gemini{}. In both setups, \gpt{} was used as the \textit{LLM-Translator}. The experimental parameters are summarized in Table~\ref{tab:exp-parameters}.

\subsection{Correctness criteria and evaluation}

In evaluating correctness, we assume that both actions in the considered games are valid and can be justified by factors such as risk aversion or cooperativeness (e.g., a cooperatively minded agent may choose “Cooperate” even if it does not yield the highest utility). A reasoning sample is marked as incorrect only if it contains logical or mathematical errors related to payoff assignment or inference—errors that human participants typically do not make (e.g., “payoff 3 is higher than payoff 5”). Misuse of game-theoretic terminology (e.g., incorrect application of Nash equilibrium) does not count against correctness. Accordingly, the evaluation focuses strictly on the logical accuracy of reasoning about payoffs, independent of theoretical knowledge.

To assess the framework’s accuracy in detecting such errors, we conducted a manual evaluation of all reasoning samples generated in the experiments. To ensure objectivity and reproducibility, we developed an evaluation protocol and accompanying scoring sheet (available in the repository). Three independent evaluators reviewed each sample to reduce the impact of individual oversight. Due to the task’s cognitive demands, some errors may have been missed, and false negatives (missed errors) are more likely than false positives. Thus, we adopted a conservative aggregation method, labeling a sample as correct only if all evaluators agreed on its correctness.

Additionally, to assess the utility of the LLM-as-a-judge~\cite{zheng2023judging} approach, we evaluated the same reasoning samples using two LLMs: \gpt{} (\texttt{gpt-4o-2024-08-06} version) and Claude 3.7 Sonnet~\cite{claude37sonnet}, a state-of-the-art model not involved in generation. Each model was given a simplified version of the evaluation protocol used by human judges. Their ratings were then compared against human evaluations.

\section{Results}
\label{sec:results}

\subsection{Experiments}

\subsubsection{Attempts distribution}
\begin{table}[htb]
\caption{Distribution of reasoning for \gpt{} and \gemini{}.}
\centering
\begin{tabular}{lccccc}
 \toprule
\textbf{Attempts} & \textbf{1} & \textbf{2} & \textbf{3} & \textbf{4} & \textbf{5} \\ 
\midrule
\gpt{} & 24 & 46 & 10 & 6 & ~4 \\
\gemini{} & 26 & 20 & 11 & 9 & 24 \\
 \bottomrule
\end{tabular}
\label{tab:attempt-dist}
\end{table}

Table~\ref{tab:attempt-dist} presents the distribution of reasoning attempts for \gpt{} and \gemini{}. For \gpt{}, 78\% of samples required only one or two attempts, with the maximum number of attempts (five) reached in just four cases. In all four instances, the final attempt was correct. In contrast, \gemini{} performed worse: 27\% of its samples reached the five-attempt limit, and in 83\% of those cases, the final attempt remained incorrect.

\subsubsection{Choices distribution}

\begin{table}
\centering
\caption{Choice `B' distribution for initial and final reasoning (\%).}
\begin{tabular}{l S[table-format=3.2] S[table-format=3.2] S[table-format=3.2] S[table-format=3.2]}
\toprule
& \multicolumn{2}{c}{\gpt{}} & \multicolumn{2}{c}{\gemini{}} \\ 
\cmidrule(lr){2-3} \cmidrule(lr){4-5}
Game & {Initial (\%)} & {Final (\%)} & {Initial (\%)} & {Final (\%)}\\
\midrule
HD & 100.00 & 100.00 & 66.67 & 70.00 \\
PD & 96.67 & 23.33 & 36.67 & 30.00 \\
SH & 100.00 & 33.33 & 80.00 & 60.00 \\
\bottomrule
\end{tabular}
\label{tab:choices-dist}
\end{table}

Table~\ref{tab:choices-dist} shows the frequency of choosing action `B' in both the initial and final reasoning attempts for \gpt{} and \gemini{}. When no feedback was provided, the initial choice remained unchanged. The action `B' corresponds to `Dove' in Hawk-Dove, `Cooperate' in the Prisoner's Dilemma, and `Stag' in Stag Hunt.

Overall, risk-averse choices became more frequent in final attempts. The Hawk-Dove game showed the highest consistency, with `Dove' being the dominant action for both models: \gpt{} selected it in 100\% of initial and final attempts, while \gemini{} increased from 66.67\% initially to 70\% in the final attempt. This reflects a preference for avoiding the worst possible payoff (0), even at the expense of forgoing the highest payoff.

In contrast, the Prisoner’s Dilemma exhibited the most pronounced shift. For \gpt{}, the frequency of choosing `Cooperate' dropped from 96.67\% to 23.33\% after feedback, indicating that initial cooperative choices were often based on incorrect reasoning. In the Stag Hunt, feedback similarly led to reduced selection of the less risk-averse `Stag': for \gpt{}, from 100\% to 33\%, and for \gemini{}, from 80\% to 60\%.

\subsubsection{Correctness}
\label{sec:correctness}

\begin{table}[t]
   \centering
   \caption{Percentage of correct reasoning samples in the original and final attempt.}
\begin{tabular}{l S[table-format=2.2] S[table-format=2.2] S[table-format=2.2] S[table-format=2.2]}
\toprule
& \multicolumn{2}{c}{\gpt{}} & \multicolumn{2}{c}{\gemini{}} \\ \cmidrule(lr){2-3} \cmidrule(lr){4-5}
Game & {Initial (\%)} & {Final (\%)} & {Initial (\%)} & {Final (\%)} \\
\midrule
HD & 46.67 & 73.33 & 25.81 & 35.48 \\
PD & 3.33 & 70.00 & 17.24 & 34.48 \\
SH & 16.67 & 90.00 & 13.33 & 40.00 \\
\bottomrule
\end{tabular}
\label{tab:cor_perc}
\end{table}

Table~\ref{tab:cor_perc} presents the percentage of correct reasoning samples in the initial and final attempts for both models. Initially, both LLMs exhibited low correctness: \gpt{} achieved an average of 22.22\%, while \gemini{} achieved 18.79\%. After receiving corrective feedback, \gpt{} demonstrated significant improvement, reaching an average correctness of 77.78\%, whereas \gemini{} showed lower gain, reaching 36.65\%.

A detailed analysis of the reasoning samples revealed that \gpt{} was often able to incorporate feedback effectively, correcting prior errors—sometimes resulting in a change of action. This effect was particularly notable in the Prisoner's Dilemma and Stag Hunt scenarios (see Table~\ref{tab:choices-dist}). In contrast, \gemini{}'s reasoning frequently reiterated previous justifications, either repeating the same errors or introducing new ones. This pattern is consistent with the number of required attempts reported in Table~\ref{tab:attempt-dist}.

\subsection{Evaluation}

\subsubsection{Confusion matrices and accuracy}
\label{sec:confusion_matrix}

\begin{table}[htb]
\caption{Confusion matrices for initial reasoning for \gemini{}. Rows: actual correctness, columns: predicted correctness, 1: correct, 0: incorrect, A: accuracy.}
\centering
\label{tab:conf_matrices_gemini}

\resizebox{\columnwidth}{!}{
\footnotesize
\begin{tabular}{@{\hskip 1em}c@{\hskip 1em}c@{\hskip 1em}c@{\hskip 1em}c@{\hskip 1em}}
\textbf{Total (A=86\%)} & \textbf{HD (A=81\%)} & \textbf{PD (A=86\%)} & \textbf{SH (A=90\%)} \\
\begin{tabular}{ccc}
\toprule
& \textbf{1} & \textbf{0} \\
\midrule
\textbf{1} & 15 & 2 \\
\textbf{0} & 11 & 62 \\
\bottomrule
\end{tabular}
&
\begin{tabular}{ccc}
\toprule
& \textbf{1} & \textbf{0} \\
\midrule
\textbf{1} & 8 & 6 \\
\textbf{0} & 0 & 17 \\
\bottomrule
\end{tabular}
&
\begin{tabular}{ccc}
\toprule
& \textbf{1} & \textbf{0} \\
\midrule
\textbf{1} & 4 & 3 \\
\textbf{0} & 1 & 21 \\
\bottomrule
\end{tabular}
&
\begin{tabular}{ccc}
\toprule
& \textbf{1} & \textbf{0} \\
\midrule
\textbf{1} & 3 & 2 \\
\textbf{0} & 1 & 24 \\
\bottomrule
\end{tabular}
\\
\end{tabular}
}
\end{table}

\begin{table}[htb]
\caption{Confusion matrices for initial reasoning of \gpt{}. Rows: actual correctness, columns: predicted correctness, 1: correct, 0: incorrect, A: accuracy.}
\centering
\label{tab:conf_matrices_gpt4}

\resizebox{\columnwidth}{!}{
\footnotesize
\begin{tabular}{@{\hskip 1em}c@{\hskip 1em}c@{\hskip 1em}c@{\hskip 1em}c@{\hskip 1em}}
\textbf{Total (A=84\%)} & \textbf{HD (A=73\%)} & \textbf{PD (A=93\%)} & \textbf{SH (A=87\%)} \\
\begin{tabular}{ccc}
\toprule
& \textbf{1} & \textbf{0} \\
\midrule
\textbf{1} & 15 & 5 \\
\textbf{0} & 9 & 61 \\
\bottomrule
\end{tabular}
&
\begin{tabular}{ccc}
\toprule
& \textbf{1} & \textbf{0} \\
\midrule
\textbf{1} & 12 & 6 \\
\textbf{0} & 2 & 10 \\
\bottomrule
\end{tabular}
&
\begin{tabular}{ccc}
\toprule
& \textbf{1} & \textbf{0} \\
\midrule
\textbf{1} & 0 & 1 \\
\textbf{0} & 1 & 28 \\
\bottomrule
\end{tabular}
&
\begin{tabular}{ccc}
\toprule
& \textbf{1} & \textbf{0} \\
\midrule
\textbf{1} & 3 & 2 \\
\textbf{0} & 2 & 23 \\
\bottomrule
\end{tabular}
\\
\end{tabular}
}
\end{table}

Confusion matrices were computed by comparing manual evaluations (representing ground-truth correctness) with the presence or absence of failed predicates (representing predicted correctness). Tables~\ref{tab:conf_matrices_gemini} and~\ref{tab:conf_matrices_gpt4} show the confusion matrices for the initial reasoning attempts of both LLMs. Overall, error recognition was high, with accuracy rates of 84\% for \gpt{} and 86\% for \gemini{}. The lowest detection accuracy was observed in the Hawk-Dove game---73\% for \gpt{} and 81\% for \gemini{}. Among the detection errors, false positives were more common than false negatives.

False positives undermine trustworthiness, as they indicate correct reasoning being incorrectly flagged as erroneous. In contrast, false negatives primarily increase computational cost without compromising correctness. Since the feedback generated by the solver is always factually accurate, false negatives are unlikely to cause an agent to revise correct reasoning into incorrect reasoning.

\subsubsection{Inter-rater agreement}

\begin{figure}
    \centering
    \includegraphics[width=0.65\linewidth]{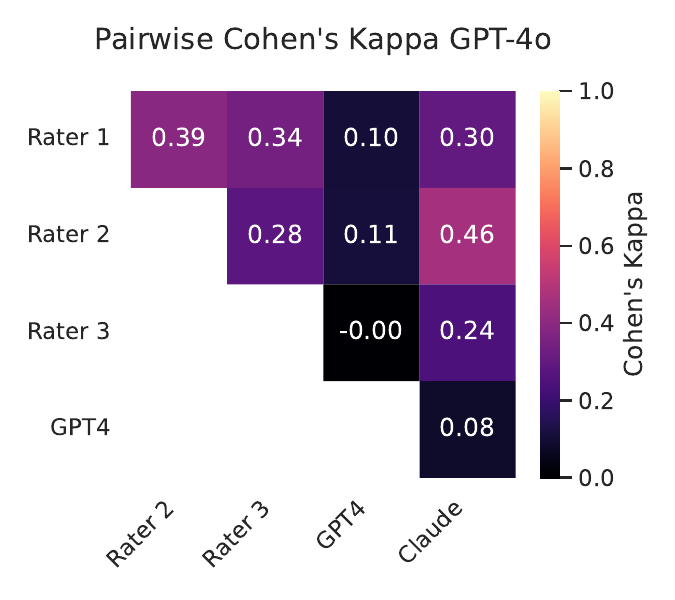}
    \caption{Pairwise inter-rater agreement for \gpt{}.}
    \label{fig:inter-rater-gpt}
\end{figure}

\begin{figure}[htbp]
    \centering
    \includegraphics[width=0.65\linewidth]{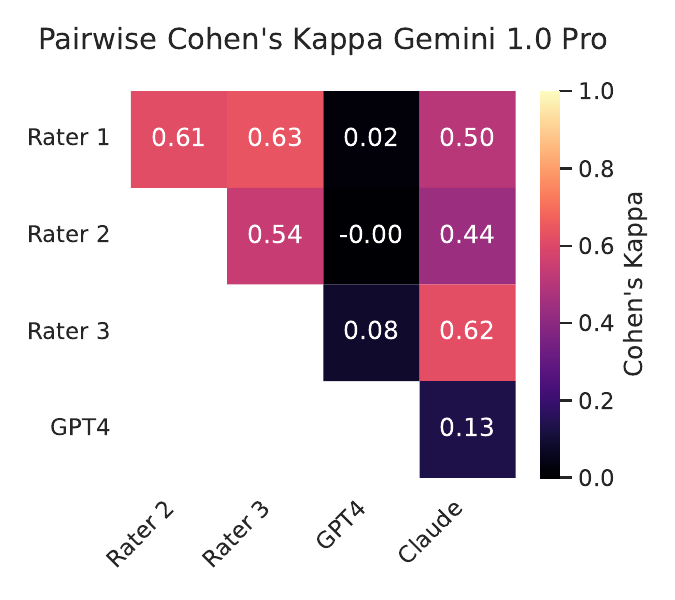}
    \caption{Pairwise inter-rater agreement for \gemini{}.}
    \label{fig:inter-rater-gemini}
\end{figure}

To assess inter-rater agreement among the three human evaluators, Fleiss’ Kappa~\cite{fleiss1971measuring} was computed. For \gpt{}, the Kappa value was $\kappa=0.31$, indicating fair agreement, while for \gemini{}, it was $\kappa=0.59$, indicating moderate agreement. These results underscore the inherent challenges of the evaluation task. The higher agreement for \gemini{} likely reflects the presence of more apparent and easier-to-detect errors compared to those found in \gpt{}.

To explore LLM-as-a-judge as a scalable alternative to human evaluation, we measured agreement using Cohen's Kappa~\cite{cohen1960coefficient} between each LLM and the aggregated human judgment, as well as pairwise agreement among all raters (human and LLMs). Figures~\ref{fig:inter-rater-gpt} and~\ref{fig:inter-rater-gemini} display the pairwise Cohen’s Kappa values for reasoning generated by \gemini{} and \gpt{}, respectively.

\gpt{}, when acting as a judge, showed low agreement with both human raters and Claude 3.7 Sonnet. In contrast, Claude 3.7 Sonnet exhibited substantially higher agreement with human raters—comparable to inter-human agreement.

For reasoning generated by \gemini{}, the agreement between \gpt{} as a judge and the aggregated human judgment was low, with Cohen’s Kappa \( \kappa = 0.044 \). In contrast, Claude 3.7 Sonnet showed substantially higher agreement with the aggregated human ratings, achieving \( \kappa = 0.555 \). For reasoning generated by \gpt{}, the agreement between \gpt{} as a judge and the aggregated human judgment was similarly low, with \( \kappa = 0.159 \). Claude 3.7 Sonnet again showed better alignment, with \( \kappa = 0.301 \).

\subsubsection{Computational cost}

In experiments with \gpt{}, an average of 1269.77 tokens were used per attempt in the instruction prompt, and the model generated an average of 1224.38 tokens in response. The average duration of a single reasoning attempt was 16.53 seconds for \gpt{} and 34.66 seconds for \gemini{}. The total runtime for all experiments was approximately 53 minutes for \gpt{} and 173 minutes for Gemini.

\section{Discussion}
\label{sec:discussion}

\paragraph{LELMA Accuracy}
The framework achieved 84\% accuracy in detecting reasoning correctness for \gpt{} and 86\% for \gemini{}. Although error detection accuracy was high for both models, the improvement resulting from corrective feedback was considerably greater for \gpt{} (ranging from 27 to 73 percentage points) compared to \gemini{} (10 to 27 percentage points). Notably, after receiving validation-based feedback, \gpt{}'s cooperation rate in the Prisoner’s Dilemma (PD) aligned more closely with rates observed in human participant experiments~\cite{bo2005cooperation}. \gpt{} more effectively incorporated feedback to refine its reasoning. By contrast, while \gemini{} was often less successful at correcting errors, it occasionally produced errors related to incorrect calculations of expected payoffs—issues not covered by the existing set of predicates. This suggests that predicate design should be preceded by systematic error sampling to ensure adequate coverage of model-specific error types.

\paragraph{LLM Performance in Social Dilemmas}
LLMs have attracted increasing interest as potential agents in game-theoretic contexts~\cite{akata2023playing,fan2023can,guo2023gpt,lore2023strategic}. This study evaluated the correctness of strategic reasoning produced by \gpt{} and \gemini{} across three one-shot games. Initial correctness, prior to any feedback, was generally low, ranging from 3.33\% to 46.67\%. These results highlight the limitations of existing automated benchmarks (e.g.,~\cite{duan2024gtbench}), which tend to emphasize outcome optimality rather than the validity of reasoning. Furthermore, the findings underscore the challenges of employing LLMs as agents in social simulations.

Despite the high overall accuracy in detecting incorrect reasoning, the process was not entirely error-free. Detection performance depended on the accuracy of translating natural language into formal predicates, a task handled by a separate LLM instance. Although translating atomic statements into formal representations is generally simpler than generating coherent reasoning, it remains a nontrivial challenge with room for improvement. One potential avenue for enhancing translation — and thus improving error detection — is to fine-tune the LLM responsible for the translation task.

\paragraph{LELMA Evaluation}
The moderate and fair inter-rater agreement values indicate that this evaluation task is challenging even for human judges. The lower agreement observed for the more advanced model suggests that in open-ended tasks such as this, more sophisticated outputs demand greater cognitive effort to assess, making errors harder to detect. Although manual evaluation remains essential at this stage, it is not a scalable solution. Potential alternatives include automated and semi-automated methods, such as LLM-derived metrics, prompting LLMs, human-LLM collaborative evaluation, and fine-tuning approaches~\cite{gao2024llm}.

We assessed the LLM-as-a-judge approach using an adapted version of our evaluation protocol. Among the two tested models, Claude 3.7 Sonnet achieved inter-rater agreement levels comparable to those between human judges. However, in the more challenging case of reasoning generated by \gpt{}, the agreement between Claude and aggregated human evaluations was low, indicating frequent omission of errors. Thus, it cannot yet be considered a reliable alternative to human judgment in this context. Among the remaining options, fine-tuning appears particularly promising, provided that a sufficiently large and representative training dataset can be gathered.

\paragraph{Limitations} This study has several limitations. First, the need for manual assessment of reasoning correctness constrained the size of the evaluation sample. Second, the accuracy of translating natural language into formal representations directly impacts the effectiveness of reasoning error detection. Imperfect translation may lead to false negatives, which undermine trustworthiness, and false positives, which increase computational cost and may inadvertently bias the model toward specific choices. Therefore, further improvement in translation accuracy is needed.

\section{Conclusions and Future Work}
\label{sec:conclusions}

This work introduces LELMA, a framework for validating the logical soundness of natural language reasoning generated by large language models.
The initial evaluation of \gpt{} and \gemini{} in game-theoretic scenarios revealed frequent reasoning errors, particularly in payoff attribution and comparison. Experimental results demonstrated that reasoning correctness improved for both models—especially for \gpt{}—when integrated within the LELMA framework. The evaluation also highlighted the need to improve translation accuracy and underscored the challenges of assessing ambiguous, open-ended tasks.

Future work will focus on enhancing the accuracy of natural language to query translation, potentially through fine-tuning. We also plan to validate the framework across a wider range of language models. Given LELMA's modular design, which enables the flexible substitution of LLM agents, query sets, and solvers, we aim to extend its application to additional domains beyond game theory.

\begin{ack}
This work was supported by a Leverhulme Trust International Professorship Grant (LIP-2022-001). We also thank Heartwin Haveluck for his assistance in the evaluation.
\end{ack}

\bibliography{references}

\begin{thebibliography}{41}
\providecommand{\natexlab}[1]{#1}
\providecommand{\url}[1]{\texttt{#1}}
\expandafter\ifx\csname urlstyle\endcsname\relax
  \providecommand{\doi}[1]{doi: #1}\else
  \providecommand{\doi}{doi: \begingroup \urlstyle{rm}\Url}\fi

\bibitem[Achiam et~al.(2023)Achiam, Adler, Agarwal, et~al.]{achiam2023gpt}
J.~Achiam, S.~Adler, S.~Agarwal, et~al.
\newblock Gpt-4 technical report.
\newblock \emph{arXiv preprint arXiv:2303.08774}, 2023.

\bibitem[Akata et~al.(2023)Akata, Schulz, Coda-Forno, et~al.]{akata2023playing}
E.~Akata, L.~Schulz, J.~Coda-Forno, et~al.
\newblock Playing repeated games with large language models.
\newblock \emph{arXiv preprint arXiv:2305.16867}, 2023.

\bibitem[Anthropic(2025)]{claude37sonnet}
Anthropic.
\newblock Claude 3.7 sonnet.
\newblock \url{https://www.anthropic.com/claude}, 2025.
\newblock Large language model.

\bibitem[Anwar et~al.(2024)Anwar, Saparov, Rando, Paleka, Turpin, Hase, Lubana,
  Jenner, Casper, Sourbut, et~al.]{anwar2024foundational}
U.~Anwar, A.~Saparov, J.~Rando, D.~Paleka, M.~Turpin, P.~Hase, E.~S. Lubana,
  E.~Jenner, S.~Casper, O.~Sourbut, et~al.
\newblock Foundational challenges in assuring alignment and safety of large
  language models.
\newblock \emph{arXiv preprint arXiv:2404.09932}, 2024.

\bibitem[B{\'o}(2005)]{bo2005cooperation}
P.~D. B{\'o}.
\newblock Cooperation under the shadow of the future: experimental evidence
  from infinitely repeated games.
\newblock \emph{American economic review}, 95\penalty0 (5):\penalty0
  1591--1604, 2005.

\bibitem[Chen et~al.(2023)Chen, Gandhi, Zhang, and Fan]{chen2023nl2tl}
Y.~Chen, R.~Gandhi, Y.~Zhang, and C.~Fan.
\newblock Nl2tl: Transforming natural languages to temporal logics using large
  language models.
\newblock In \emph{Proceedings of the 2023 Conference on Empirical Methods in
  Natural Language Processing}, pages 15880--15903, 2023.

\bibitem[Cohen(1960)]{cohen1960coefficient}
J.~Cohen.
\newblock A coefficient of agreement for nominal scales.
\newblock \emph{Educational and psychological measurement}, 20\penalty0
  (1):\penalty0 37--46, 1960.

\bibitem[Cosler et~al.(2023)Cosler, Hahn, Mendoza, Schmitt, and
  Trippel]{cosler2023nl2spec}
M.~Cosler, C.~Hahn, D.~Mendoza, F.~Schmitt, and C.~Trippel.
\newblock nl2spec: interactively translating unstructured natural language to
  temporal logics with large language models.
\newblock In \emph{International Conference on Computer Aided Verification},
  pages 383--396. Springer, 2023.

\bibitem[Duan et~al.(2024)Duan, Zhang, Diffenderfer, Kailkhura, Sun,
  Stengel-Eskin, Bansal, Chen, and Xu]{duan2024gtbench}
J.~Duan, R.~Zhang, J.~Diffenderfer, B.~Kailkhura, L.~Sun, E.~Stengel-Eskin,
  M.~Bansal, T.~Chen, and K.~Xu.
\newblock Gtbench: Uncovering the strategic reasoning limitations of llms via
  game-theoretic evaluations.
\newblock \emph{arXiv preprint arXiv:2402.12348}, 2024.

\bibitem[Durante et~al.(2024)Durante, Huang, Wake, Gong, Park, Sarkar, Taori,
  Noda, Terzopoulos, Choi, et~al.]{durante2024agent}
Z.~Durante, Q.~Huang, N.~Wake, R.~Gong, J.~S. Park, B.~Sarkar, R.~Taori,
  Y.~Noda, D.~Terzopoulos, Y.~Choi, et~al.
\newblock Agent ai: Surveying the horizons of multimodal interaction.
\newblock \emph{arXiv preprint arXiv:2401.03568}, 2024.

\bibitem[Fan et~al.(2024)Fan, Chen, Jin, and He]{fan2023can}
C.~Fan, J.~Chen, Y.~Jin, and H.~He.
\newblock Can large language models serve as rational players in game theory? a
  systematic analysis.
\newblock In \emph{Proceedings of the AAAI Conference on Artificial
  Intelligence}, volume~38, pages 17960--17967, 2024.

\bibitem[Feng et~al.(2023)Feng, Xu, Hao, Sharma, Shen, Zhao, and
  Chen]{feng2023language}
J.~Feng, R.~Xu, J.~Hao, H.~Sharma, Y.~Shen, D.~Zhao, and W.~Chen.
\newblock Language models can be logical solvers.
\newblock \emph{arXiv preprint arXiv:2311.06158}, 2023.

\bibitem[Fleiss(1971)]{fleiss1971measuring}
J.~L. Fleiss.
\newblock Measuring nominal scale agreement among many raters.
\newblock \emph{Psychological bulletin}, 76\penalty0 (5):\penalty0 378, 1971.

\bibitem[Gao et~al.(2024)Gao, Hu, Ruan, Pu, and Wan]{gao2024llm}
M.~Gao, X.~Hu, J.~Ruan, X.~Pu, and X.~Wan.
\newblock {LLM}-based nlg evaluation: Current status and challenges.
\newblock \emph{arXiv preprint arXiv:2402.01383}, 2024.

\bibitem[Genesereth et~al.(2005)Genesereth, Love, and Pell]{gdl}
M.~R. Genesereth, N.~Love, and B.~Pell.
\newblock General game playing: Overview of the {AAAI} competition.
\newblock \emph{{AI} Mag.}, 26\penalty0 (2):\penalty0 62--72, 2005.

\bibitem[Giacomo et~al.(2010)Giacomo, Lesp{\'{e}}rance, and Pearce]{sc-games1}
G.~D. Giacomo, Y.~Lesp{\'{e}}rance, and A.~R. Pearce.
\newblock Situation calculus based programs for representing and reasoning
  about game structures.
\newblock In F.~Lin, U.~Sattler, and M.~Truszczynski, editors, \emph{Principles
  of Knowledge Representation and Reasoning: Proceedings of the Twelfth
  International Conference, {KR} 2010, Toronto, Ontario, Canada, May 9-13,
  2010}. {AAAI} Press, 2010.

\bibitem[Guo(2023)]{guo2023gpt}
F.~Guo.
\newblock {GPT} agents in game theory experiments.
\newblock \emph{arXiv preprint arXiv:2305.05516}, 2023.

\bibitem[He-Yueya et~al.(2023)He-Yueya, Poesia, Wang, and
  Goodman]{he2023solving}
J.~He-Yueya, G.~Poesia, R.~E. Wang, and N.~D. Goodman.
\newblock Solving math word problems by combining language models with symbolic
  solvers.
\newblock \emph{arXiv preprint arXiv:2304.09102}, 2023.

\bibitem[Imani et~al.(2023)Imani, Du, and Shrivastava]{imani2023mathprompter}
S.~Imani, L.~Du, and H.~Shrivastava.
\newblock Mathprompter: Mathematical reasoning using large language models.
\newblock \emph{arXiv preprint arXiv:2303.05398}, 2023.

\bibitem[Jiang et~al.(2022)Jiang, Welleck, Zhou, Li, Liu, Jamnik, Lacroix, Wu,
  and Lample]{jiang2022draft}
A.~Q. Jiang, S.~Welleck, J.~P. Zhou, W.~Li, J.~Liu, M.~Jamnik, T.~Lacroix,
  Y.~Wu, and G.~Lample.
\newblock Draft, sketch, and prove: Guiding formal theorem provers with
  informal proofs.
\newblock \emph{arXiv preprint arXiv:2210.12283}, 2022.

\bibitem[Kambhampati et~al.(2024)Kambhampati, Valmeekam, Guan, Verma, Stechly,
  Bhambri, Saldyt, and Murthy]{kambhampati2024llms}
S.~Kambhampati, K.~Valmeekam, L.~Guan, M.~Verma, K.~Stechly, S.~Bhambri,
  L.~Saldyt, and A.~Murthy.
\newblock Llms can't plan, but can help planning in llm-modulo frameworks.
\newblock \emph{arXiv preprint arXiv:2402.01817}, 2024.

\bibitem[K{\"{o}}rner et~al.(2022)K{\"{o}}rner, Leuschel, Barbosa, Costa, Dahl,
  Hermenegildo, Morales, Wielemaker, Diaz, Abreu, and Ciatto]{p50}
P.~K{\"{o}}rner, M.~Leuschel, J.~Barbosa, V.~S. Costa, V.~Dahl, M.~V.
  Hermenegildo, J.~F. Morales, J.~Wielemaker, D.~Diaz, S.~Abreu, and G.~Ciatto.
\newblock Fifty {Y}ears of {P}rolog and {B}eyond.
\newblock \emph{Theory and Practice of Logic Programming}, 22\penalty0
  (6):\penalty0 776–858, 2022.
\newblock \doi{10.1017/S1471068422000102}.

\bibitem[Lesperance et~al.(2024)Lesperance, De~Giacomo, Rostamigiv, and
  Khan]{sc-games2}
Y.~Lesperance, G.~De~Giacomo, M.~Rostamigiv, and S.~M. Khan.
\newblock Abstraction of situation calculus concurrent game structures.
\newblock \emph{Proceedings of the AAAI Conference on Artificial Intelligence},
  38\penalty0 (9):\penalty0 10624--10634, Mar. 2024.
\newblock \doi{10.1609/aaai.v38i9.28933}.
\newblock URL \url{https://ojs.aaai.org/index.php/AAAI/article/view/28933}.

\bibitem[Lor{\`e} and Heydari(2023)]{lore2023strategic}
N.~Lor{\`e} and B.~Heydari.
\newblock Strategic behavior of large language models: Game structure vs.
  contextual framing.
\newblock \emph{arXiv preprint arXiv:2309.05898}, 2023.

\bibitem[Love et~al.(2006)Love, Hinrichs, Haley, Schkufza, and
  Genesereth]{gdl-orig}
N.~Love, T.~Hinrichs, D.~Haley, E.~Schkufza, and M.~Genesereth.
\newblock {General Game Playing: Game Description Language Specification}.
\newblock Technical Report LG–2006–01, Stanford University, 2006.

\bibitem[McCarthy and Hayes(1981)]{sc}
J.~McCarthy and P.~Hayes.
\newblock Some philosophical problems from the standpoint of artificial
  intelligence.
\newblock In B.~L. Webber and N.~J. Nilsson, editors, \emph{Readings in
  Artificial Intelligence}, pages 431--450. Morgan Kaufmann, 1981.

\bibitem[{OpenAI}(2024)]{gpt4o2024}
{OpenAI}.
\newblock {GPT}-4o.
\newblock \url{https://openai.com/index/hello-gpt-4o/}, 2024.
\newblock URL \url{https://openai.com/index/hello-gpt-4o/}.
\newblock Accessed: 25/07/2024.

\bibitem[Osborne(2004)]{Osborne2004}
M.~Osborne.
\newblock \emph{Introduction to Game Theory}.
\newblock Oxford University Press USA, 2004.

\bibitem[Pan et~al.(2023)Pan, Albalak, Wang, and Wang]{pan2023logic}
L.~Pan, A.~Albalak, X.~Wang, and W.~Wang.
\newblock Logic-lm: Empowering large language models with symbolic solvers for
  faithful logical reasoning.
\newblock In \emph{Findings of the Association for Computational Linguistics:
  EMNLP 2023}, pages 3806--3824, 2023.

\bibitem[Park et~al.(2023)Park, O'Brien, Cai, Morris, Liang, and
  Bernstein]{park2023generative}
J.~S. Park, J.~O'Brien, C.~J. Cai, M.~R. Morris, P.~Liang, and M.~S. Bernstein.
\newblock Generative agents: Interactive simulacra of human behavior.
\newblock In \emph{Proceedings of the 36th annual acm symposium on user
  interface software and technology}, pages 1--22, 2023.

\bibitem[Rasmusen(2006)]{Rasmusen2004introtogametheory}
E.~Rasmusen.
\newblock \emph{Games and information an introduction to game theory}.
\newblock Blackwell, 2006.

\bibitem[Rohan~Anil et~al.(2024)]{geminiteam2024gemini}
J.-B.~A. Rohan~Anil, Sebastian~Borgeaud et~al.
\newblock Gemini: A family of highly capable multimodal models.
\newblock \emph{arXiv preprint arXiv:2312.11805}, 2024.

\bibitem[Scherl and Levesque(2003)]{sc-kbf}
R.~B. Scherl and H.~J. Levesque.
\newblock Knowledge, action, and the frame problem.
\newblock \emph{Artif. Intell.}, 144\penalty0 (1-2):\penalty0 1--39, 2003.
\newblock \doi{10.1016/S0004-3702(02)00365-X}.
\newblock URL \url{https://doi.org/10.1016/S0004-3702(02)00365-X}.

\bibitem[Schiffel and Thielscher(2011)]{Schiffel_Thielscher_2011}
S.~Schiffel and M.~Thielscher.
\newblock Reasoning about general games described in gdl-ii.
\newblock \emph{Proceedings of the AAAI Conference on Artificial Intelligence},
  25\penalty0 (1):\penalty0 846--851, Aug. 2011.
\newblock \doi{10.1609/aaai.v25i1.7944}.

\bibitem[Thielscher(2010)]{gdl-ii}
M.~Thielscher.
\newblock A general game description language for incomplete information games.
\newblock In \emph{Proceedings of the Twenty-Fourth AAAI Conference on
  Artificial Intelligence}, AAAI'10, page 994–999. AAAI Press, 2010.

\bibitem[Thielscher(2011)]{gdl-univ}
M.~Thielscher.
\newblock The general game playing description language is universal.
\newblock In T.~Walsh, editor, \emph{{IJCAI} 2011, Proceedings of the 22nd
  International Joint Conference on Artificial Intelligence, Barcelona,
  Catalonia, Spain, July 16-22, 2011}, pages 1107--1112. {IJCAI/AAAI}, 2011.

\bibitem[Wei et~al.(2022)Wei, Wang, Schuurmans, Bosma, Xia, Chi, Le, Zhou,
  et~al.]{wei2023chainofthought}
J.~Wei, X.~Wang, D.~Schuurmans, M.~Bosma, F.~Xia, E.~Chi, Q.~V. Le, D.~Zhou,
  et~al.
\newblock Chain-of-thought prompting elicits reasoning in large language
  models.
\newblock \emph{Advances in neural information processing systems},
  35:\penalty0 24824--24837, 2022.

\bibitem[Wielemaker et~al.(2012)Wielemaker, Schrijvers, Triska, and
  Lager]{wielemaker2012swi}
J.~Wielemaker, T.~Schrijvers, M.~Triska, and T.~Lager.
\newblock Swi-prolog.
\newblock \emph{Theory and Practice of Logic Programming}, 12\penalty0
  (1-2):\penalty0 67--96, 2012.

\bibitem[Wu et~al.(2022)Wu, Jiang, Li, Rabe, Staats, Jamnik, and
  Szegedy]{wu2022autoformalization}
Y.~Wu, A.~Q. Jiang, W.~Li, M.~Rabe, C.~Staats, M.~Jamnik, and C.~Szegedy.
\newblock Autoformalization with large language models.
\newblock \emph{Advances in Neural Information Processing Systems},
  35:\penalty0 32353--32368, 2022.

\bibitem[Yang et~al.(2023)Yang, Ishay, and Lee]{yang2023coupling}
Z.~Yang, A.~Ishay, and J.~Lee.
\newblock Coupling large language models with logic programming for robust and
  general reasoning from text.
\newblock \emph{arXiv preprint arXiv:2307.07696}, 2023.

\bibitem[Zheng et~al.(2023)Zheng, Chiang, Sheng, Zhuang, Wu, Zhuang, Lin, Li,
  Li, Xing, et~al.]{zheng2023judging}
L.~Zheng, W.-L. Chiang, Y.~Sheng, S.~Zhuang, Z.~Wu, Y.~Zhuang, Z.~Lin, Z.~Li,
  D.~Li, E.~Xing, et~al.
\newblock Judging llm-as-a-judge with mt-bench and chatbot arena.
\newblock \emph{Advances in Neural Information Processing Systems},
  36:\penalty0 46595--46623, 2023.

\end{thebibliography}


\end{document}